\documentclass[preprint,12pt]{elsarticle}
\usepackage{amssymb}
\usepackage{amsmath}
\usepackage[hidelinks]{hyperref}
\usepackage{lineno}
\usepackage{booktabs}
\usepackage{placeins}
\usepackage{float}
\usepackage{siunitx} 
\usepackage[capitalise,noabbrev]{cleveref}
\usepackage{subcaption}

\begin{document}
\begin{frontmatter}

\title{Implicit neural representations for larval zebrafish brain microscopy: a reproducible benchmark on the MapZebrain atlas}

\author{Agnieszka Pregowska} 

\affiliation{organization={Institute of Fundamental Technological Research
Polish Academy of Sciences},
            addressline={Pawinskiego 5B}, 
            city={Warsaw},
            postcode={02-106}, 
            country={Poland}}
\begin{abstract}
\textbf{Background:} 
Implicit neural representations (INRs) provide continuous, coordinate-based encodings that are attractive for atlas registration, cross-modality resampling, sparse-view completion, and compact sharing of neuroanatomical datasets. Reproducible evaluation is missing for high-resolution larval zebrafish microscopy, where accurate neuropil delineation and preservation of fine neuronal processes are critical.

\textbf{New method:} 
We introduce a reproducible INR benchmark tailored to the MapZebrain larval zebrafish brain atlas. Under a unified, seed-controlled protocol, we compare SIREN, Fourier features, Haar positional encoding, and a multi-resolution grid across 950 grayscale microscopy images, including atlas slices and single-neuron projections. Images are normalized using per-image $(1,99)$  percentiles using only $10.00\%$ of pixels from non-held-out columns. Spatial  generalization is evaluated on a deterministic $40.00\%$ column-wise hold-out along the $X$-axis.

\textbf{Results:} 
Haar and Fourier encodings achieve the highest macro-averaged reconstruction fidelity on held-out columns ($\approx 26$\,dB), with the grid representation moderately behind and SIREN trailing in macro averages but remaining competitive on area-weighted micro averages in the heterogeneous all-in-one regime. SSIM and an edge-focused error (mean absolute error on Sobel edges) confirm superior boundary preservation for Haar and Fourier.

\textbf{Comparison with existing methods:}
Compared with smooth-bias INRs such as SIREN and grid-based schemes prone to axis-aligned artefacts, explicit spectral and multiscale encodings deliver improved high-frequency fidelity and sharper inter-regional boundaries.

\textbf{Conclusions:}
For MapZebrain workflows, Haar and Fourier encodings are preferable for boundary-sensitive tasks such as atlas registration, label transfer, and morphology-preserving sharing of single-neuron projections, while lightweight SIREN remains a useful baseline for background modelling or denoising when high-frequency detail is less critical.

\end{abstract}


\end{frontmatter}


\section{Introduction}
Implicit Neural Representations (INRs) encode signals as a continuous function of spatial coordinates parameterized by a neural network. This representation supports alias-free resampling, super-resolution at any scale, and compact detail-preserving storage in imaging (\cite{sitzmann2020,tancik2020,chen2022}). INRs have reshaped graphics and view synthesis through positional encodings and radiance fields (\cite{mildenhall2022}), as well as efficient multi-resolution/grid and tensor encoders (\cite{mueller2022,fridovichkeil2022,chen2022}). Despite this progress, there is little systematic evidence comparing classical INR variants on neuroanatomical images, where signals are diffuse and island-like while boundaries remain sharp.

Different formulations of INRs show varying representational properties. Periodic-activation MLPs (SIREN) mitigate spectral bias, allowing direct modeling of high-frequency structure (\cite{sitzmann2020}). Fourier features embed input coordinates in sinusoidal bases, allowing conventional MLPs to capture fine detail efficiently in low-dimensional settings (\cite{tancik2020}). A wavelet-style Haar positional encoding offers a simple piecewise-constant alternative.

To improve computational efficiency, a recent study has explored semi-explicit encoders that map coordinates to features through multi-resolution tables. Instant Neural Graphics Primitives introduce grid encoding (\cite{mueller2022}), Plenoxels replace MLPs with dense voxel grids and spherical harmonics (\cite{fridovichkeil2022}), and TensoRF factorizes fields into low-rank tensors (\cite{chen2022}). The local Implicit Image Function further shows that local implicit decoders preserve edges and textures for arbitrary-scale super-resolution (\cite{chen2021}). Although grid- and tensor-based encoders are now standard in scene modeling, their behavior on neuroanatomical textures, characterized by sparse, island-like signals with sharp inter-regional boundaries remains insufficiently understood, motivating a head-to-head evaluation under a unified protocol (\cite{xie2022}).

In this paper, we present quantitative evaluation of INRs on MapZebrain. We study four INRs variants, including SIREN, Fourier features, Haar positional encoding, and a multi-resolution grid (Grid). We evaluate under two regimes: regions (a curated collection of atlas slices), and all-in-one (a recursive crawl that covers atlas regions and single-neuron projections), all within a unified, seed-controlled preprocessing and validation protocol. We report both macro- and pixel-weighted micro-averaged PSNR and analyze qualitative failure modes relevant to neuroanatomical textures.

To position this paper as a practical tool for larval zebrafish imaging pipelines, we introduce a reproducible INR benchmark tailored to the MapZebrain
	larval zebrafish atlas, with a unified preprocessing, column-wise hold-out, and seed-controlled train/test protocol that can be reused for future methods. We also provide a large-scale, head-to-head comparison of four classical INR 	parameterizations (SIREN, Fourier features, Haar positional encoding, and a 	multi-resolution grid) across 950 neuroanatomical microscopy images spanning 	curated atlas slices and single-neuron projections. Moreover, we relate differences in reconstruction fidelity, SSIM, and edge-focused errors to biologically relevant structures, including layered neuropil, sharp
	region boundaries, and long, thin axonal processes in the larval zebrafish brain. We derive practical recommendations for selecting INR encodings in
	atlas-centric workflows (registration, cross-modality resampling, sparse-view completion, and storage-efficient sharing of single-neuron projections) and provide code and fixed splits to enable straightforward adoption.

\section{Biological motivation}
The larval zebrafish brain is small, optically accessible, and stereotyped across individuals, with layered neuropil (e.g.\ optic tectum), sharply delineated region borders (e.g.\ pallium-diencephalon interfaces), and long, thin axonal tracts (e.g.\ commissures). These structures impose competing reconstruction demands: preserving punctate cell-body texture and synaptic-like hotspots, maintaining sharp inter-regional boundaries, and avoiding over-smoothing of submicron-scale neurites.

From a biomedical imaging standpoint, an implicit neural representation that respects these constraints is directly useful for several core downstream tasks. First, atlas registration and region delineation are inherently boundary-sensitive; fidelity at pallium-diencephalon borders and within layered neuropil translates into reduced registration error and improved label transfer. Second, cross-modality resampling, such as aligning anatomy with gene-expression volumes, requires alias-free interpolation at arbitrary spatial coordinates, a natural fit for continuous coordinate-based decoders. Third, sparse-view completion reflects practical acquisition scenarios (mosaics, partial fields of view), where spatial generalization across contiguous missing spans is critical; our column-wise hold-out explicitly probes this behavior. Finally, storage-efficient sharing of single-neuron projections benefits from compact parametric representations that preserve thin processes, commissural crossings, and local texture without inflating memory footprint.

These biological and imaging characteristics translate into concrete methodological requirements for INR models. An effective INR must preserve region borders and layering to avoid systematic registration errors, maintain thin axonal processes across held-out spans to support tracing and morphology analysis, and provide smooth, alias-free interpolation for cross-modality workflows. Positional encodings therefore become first-order design choices: multiscale or spectral schemes are expected to better preserve sharp boundaries and thin neurites, whereas models with stronger low-frequency bias risk attenuating fine morphology. Our evaluation protocol and metrics are thus chosen to reflect these practical constraints rather than generic image-denoising performance.

\section{Methods}

\subsection{Implicit Neural Representations}
Implicit representations encode a signal as a continuous function
\[
f_\theta:\mathbb{R}^d \to \mathbb{R}^c,
\]
implemented by a neural network with parameters~$\theta$. Once trained, $f_\theta$ can be queried at arbitrary coordinates, providing resolution-independent access to the underlying signal. INRs offer smooth interpolation, parameter-efficient storage, and differentiability, facilitating gradient-based pipelines (\cite{park2019,mescheder2019,mildenhall2022,strumpler2022,mueller2022}). 

\subsection{Database}
\label{sec:database}

We consider MapZebrain, a multi-modal atlas of the larval zebrafish brain integrating region-level anatomy and single-neuron projections within a common coordinate framework(\cite{randlett2015,kunst2019,shainer2023,mapzebrain_site}).
The larval zebrafish offers a small, optically accessible, and stereotyped nervous system, enabling whole-brain imaging at cellular and gene-expression resolution with high throughput. MapZebrain integrates region-level anatomy and single-neuron projections into a common coordinate frame, yielding images with punctate textures and sharp inter-regional boundaries across heterogeneous native resolution. These properties make MapZebrain a realistic and stringent benchmark for reconstruction, resampling, and compression, against which we test the capabilities and limitations of INR models. Our corpus comprises 950 images in total: 227 curated atlas slices (the regions regime) and 723 files from a recursive, heterogeneous crawl (the all-in-one regime).

\subsection{Regimes}
\label{sec:regimes}

We consider two complementary regimes. Regions: a curated set of 227 representative atlas slices (modal size $H\times W$), providing a controlled, repeatable substrate. All-in-one: a recursive crawl of 723 files comprising atlas-region images and single-neuron projections at native resolutions, reflecting real-world variability.

\subsection{Preprocessing}
\label{sec:preprocessings}

All images are converted to grayscale and normalized per image to $[0,1]$ using the $(1,99)$ percentiles:
\(I_{\mathrm{norm}}=\mathrm{clip}((I-P_1)/(P_{99}-P_1+10^{-6}),0,1)\).
In regions, when sizes differ we centrally crop to the modal size to enable grid-based batching, all-in-one preserves native sizes. Spatial generalization is probed with a column-wise hold-out along $X$ ($\alpha=0.40$). In regions, a single deterministic mask (global seed) is reused across images. In all-in-one, masks are deterministically seeded per file to avoid leakage and ensure reproducibility.

\subsection{Evaluation}
\label{sec:evaluation}

For each image, we train the INR on a random subsample of approximately 10.00\% of the available pixels, restricted to training columns, and then reconstruct the learned field $f_\theta$ over $[0,1]^2$ sampled at the original image resolution. Evaluation is performed exclusively on a column-wise hold-out along the $X$-axis ($\alpha=0.4$; \texttt{blocked\_cols\_X}), which probes spatial generalization across contiguous gaps rather than random-pixel denoising. This split is deterministic and fixed across methods to ensure comparability.

Let $I:\{0,\ldots,H-1\}\times\{0,\ldots,W-1\}\to\mathbb{R}_{\ge0}$ denote a preprocessed grayscale image normalized to $[0,1]$, and let $\hat I$ be its INR reconstruction. The test set $\Omega_{\text{test}}$ contains all pixels from held-out columns, and the reconstruction error is quantified by mean squared error
\begin{equation}
\mathrm{MSE}_{\text{test}}
=\frac{1}{|\Omega_{\text{test}}|}
\sum_{(x,y)\in\Omega_{\text{test}}}\big(I(x,y)-\hat I(x,y)\big)^2.
\end{equation}
The primary metric is the corresponding peak signal-to-noise ratio (PSNR, dB)
\begin{equation}
\mathrm{PSNR}_{\text{test}}
=20\log_{10}\!\Big(\frac{R}{\sqrt{\mathrm{MSE}_{\text{test}}}}\Big),
\qquad R=1.0,
\end{equation}
where $R$ is the dynamic range after normalization. For each method we report macro-averages (mean $\pm$ SD across images) and micro-averages (computed from a single aggregated MSE), the latter weighting larger images more heavily. Robust statistics (median, IQR, 10.00\% trimmed mean) are reported in Appendix Table \ref{tab:robu} to reduce the influence of near-constant outliers.

Performance distributions are summarized with box and violin plots. The comparison is based on per image PSNR differences. Statistical significance is assessed using the paired Wilcoxon signed-rank test. Confidence intervals for mean differences are estimated through nonparametric bootstrap with $B=10{,}000$ resamples (95.00\% percentile). We also report the number of wins per method (highest $\mathrm{PSNR}_{\text{test}}$ per image). We emphasize both macro- and micro-averages because image sizes vary widely across the corpus. Extremely high PSNR outliers in regions correspond to near-constant fields after percentile normalization. These are reported explicitly and analyzed separately. We also report effect sizes (e.g., Cliff's $\delta$ with bootstrap CIs) in addition to $p$-values to reflect practical, not only statistical, significance. 

Beyond PSNR, we compute SSIM on the held-out columns and an edge-focused error defined as the mean absolute error restricted to edge pixels identified in the ground truth by a Sobel operator. The threshold selects the top $10.00\%$ of gradient magnitudes per image, computed on the test columns only.

\section{Numerical Results}

Table \ref{tab:macro_micro} summarizes macro and micro $\mathrm{PSNR}_{\text{test}}$ across 950 images. Fourier and Haar achieve the highest macro means (both $\approx 26$\,dB), followed by Grid ($\approx 24$\,dB), whereas SIREN trails ($\approx 14$\,dB). Because a subset of atlas slices becomes near-constant after $(1,99)$ normalization, macro means substantially exceed medians (typically $12$-$14$\,dB). We therefore report robust summaries and analyze outliers separately.

\begin{table}[ht]
\centering
\caption{Macro- and micro-averaged $\mathrm{PSNR}_{\text{test\_mean}}$ (dB) across 950 images.}
\label{tab:macro_micro}
\begin{tabular}{lcccc}
\toprule
Model & Macro-Mean & Macro-Std & Micro-Mean & Wins \\
\midrule
Haar    & 26.09 & 27.70 & 12.83 & 354 \\
Fourier & 25.95 & 26.59 & 13.75 & 288 \\
SIREN   & 13.90 &  2.28 & 13.41 & 233 \\
Grid    & 24.12 & 28.31 & 10.98 &  75 \\
\bottomrule
\end{tabular}
\begin{minipage}{0.9\linewidth}
\small
\textit{Note.} Macro means are inflated by a small number of near-constant atlas slices after per-image $(1,99)$ normalization, occasionally yielding $\geq 100$\,dB PSNR. For robust statistics (median, IQR, 10.00\% trimmed mean) see \Cref{tab:robu} and the outlier analysis in \Cref{tab:outliers}.
\end{minipage}

\end{table}

Area-weighted (micro) PSNR, computed from a single MSE aggregated over all images, down-weights small atlas fields and better reflects performance on large heterogeneous projections (Table \ref{tab:micro_by_regime}). On the entire corpus, Fourier and SIREN are close on the micro average ($\approx 13.5$-$13.8$\,dB), Haar is slightly lower ($\approx 12.8$\,dB), and Grid remains behind ($\approx 11$\,dB). In the all-in-one regime, Fourier, Haar, and SIREN cluster around $12$-$13$\,dB, while Grid drops below $10$\,dB.

\begin{table}[ht]
	\centering
	\caption{Area-weighted (micro) $\mathrm{PSNR}_{\text{test}}$ (dB) by regime.}
	\label{tab:micro_by_regime}
	\begin{tabular}{lcccc}
		\toprule
		Regime & Haar & Fourier & SIREN & Grid \\
		\midrule
		all-in-one & 11.23 & 12.47 & 12.71 & 9.80 \\
		regions & 26.47 & 26.57 & 17.68 & 26.45 \\
		\bottomrule
	\end{tabular}
\end{table}

\begin{figure}[t]
    \centering
    \includegraphics[width=1\linewidth]{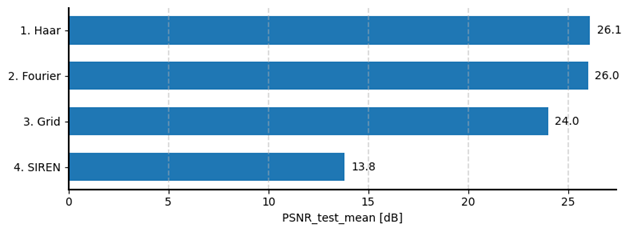}
    \caption{Mean $\mathrm{PSNR}_{\text{test\_mean}}$ (macro) across methods. Haar and Fourier achieve the highest averages.}
    \label{fig:bar_regions}
\end{figure}

\begin{figure}[t]
    \centering
    \includegraphics[width=1\linewidth]{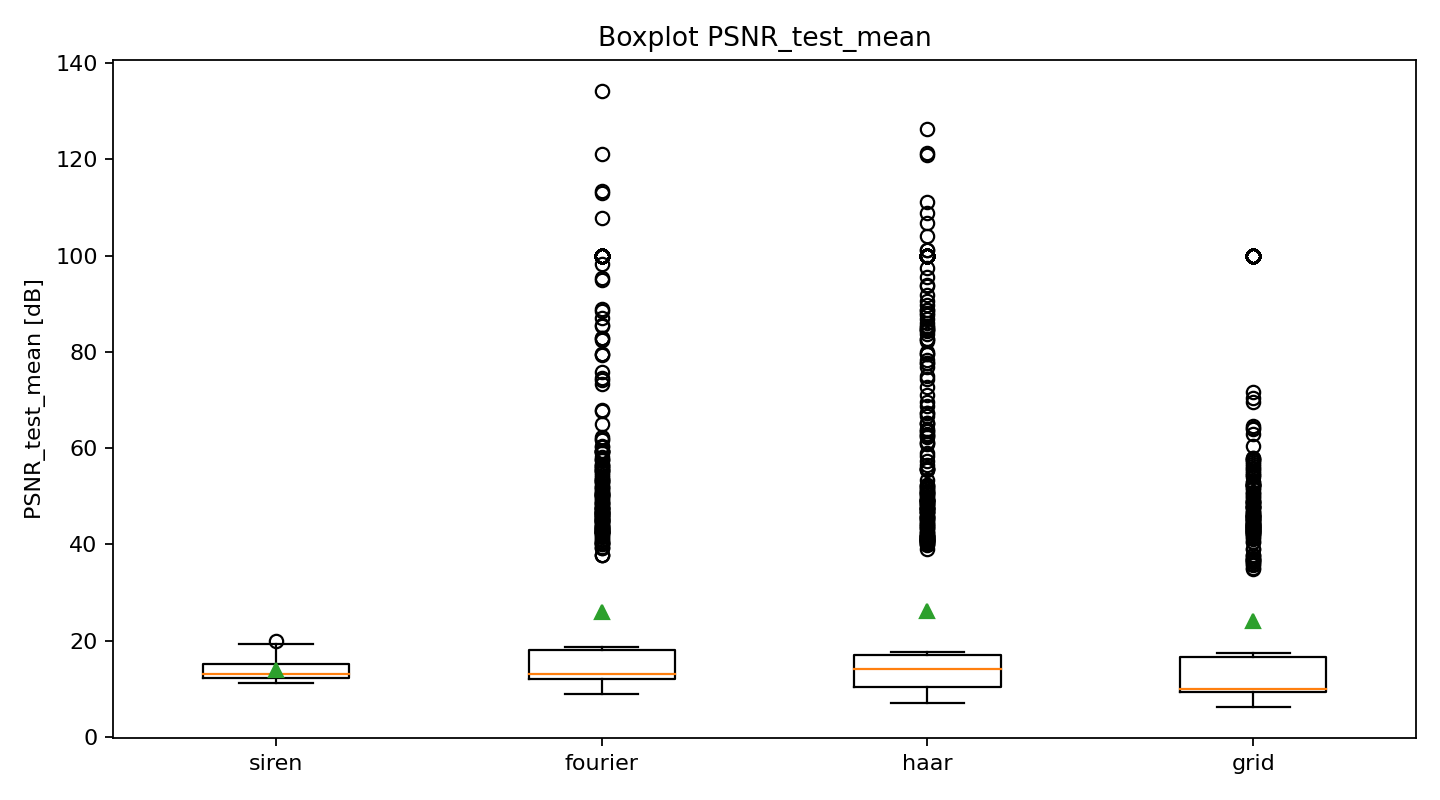}
    \caption{Box plot of $\mathrm{PSNR}_{\text{test\_mean}}$ showing broad distributions with occasional outliers.}
    \label{fig:box_regions}
\end{figure}

\begin{figure}[t]
    \centering
    \includegraphics[width=1\linewidth]{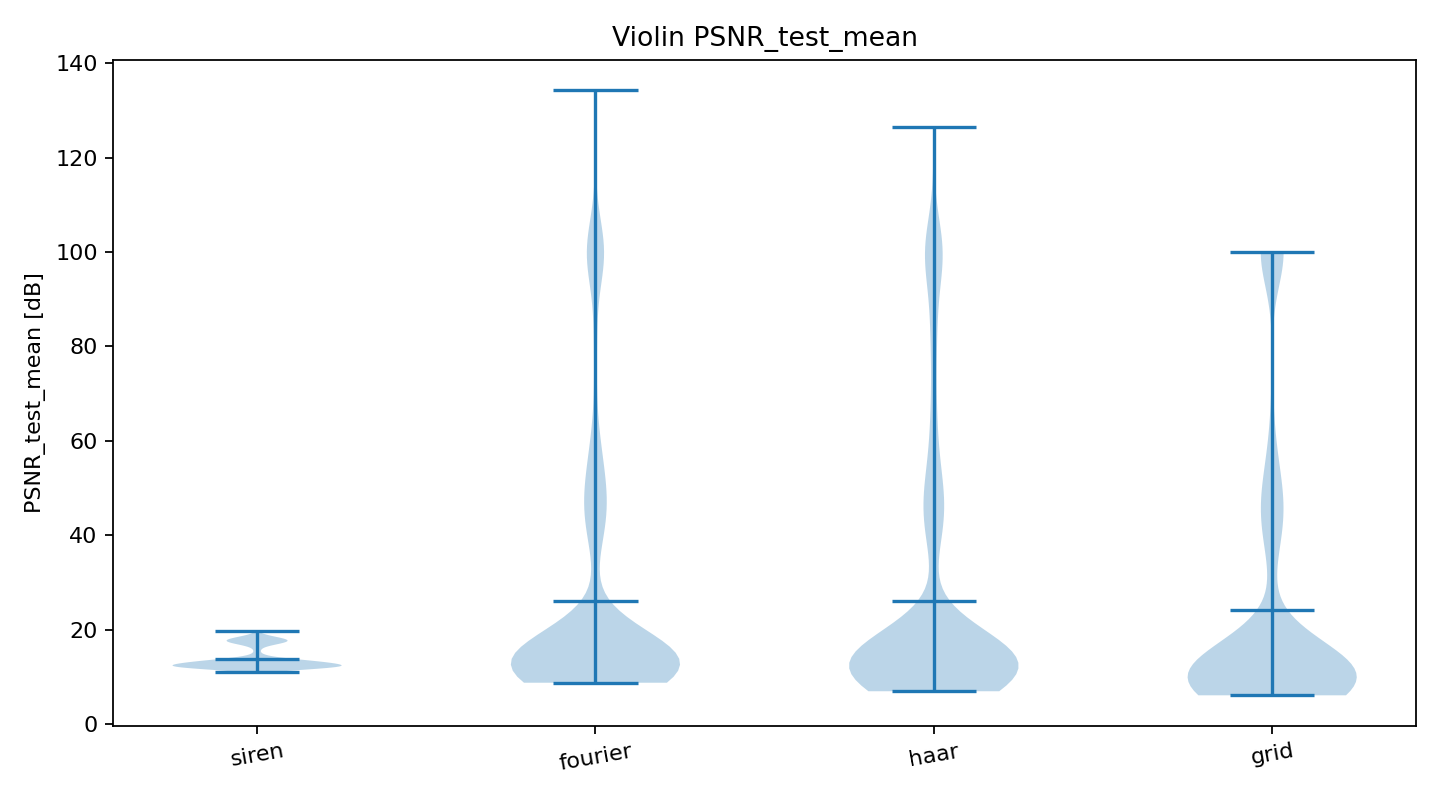}
    \caption{Violin plot of $\mathrm{PSNR}_{\text{test\_mean}}$ illustrating per-image variability.}
    \label{fig:violin_regions}
\end{figure}

Figures~\ref{fig:bar_regions}-\ref{fig:violin_regions} summarize per-image distributions (bar, box and violin).These plots summarize the same statistics as were presented in Table \ref{tab:macro_micro}. Haar and Fourier show nearly identical median PSNR distributions, with large variance caused by image heterogeneity (typical values 10-26 dB). Very high outliers (equal or higher than 100 dB) occur occasionally, mostly in low-variance regions or small images. Grid performs moderately but remains below Haar and Fourier across most samples, while SIREN maintains stable but lower scores, consistent with its known preference for low-frequency structure.

\begin{figure}[t]
	\centering
	\includegraphics[width=0.8\linewidth]{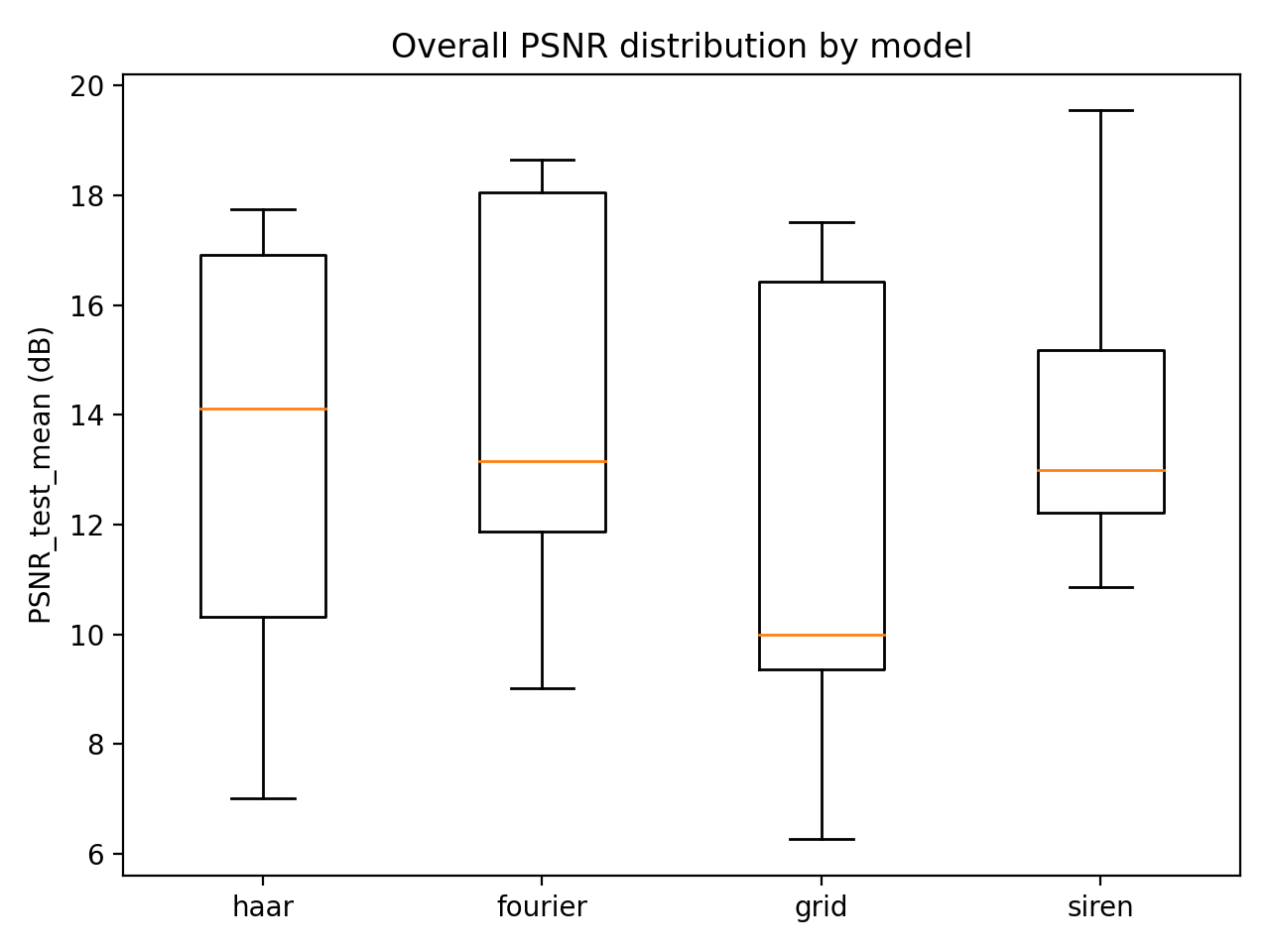}
	\caption{Overall distribution of $\mathrm{PSNR}_{\text{test\_mean}}$ for the four INR variants (all images). Fourier and Haar have similar medians but a broader upper tail. SIREN is the most concentrated.}
	\label{fig:box_overall_psnr}
\end{figure}

Figure~\ref{fig:box_overall_psnr} shows the same result in a condensed form: all four models have their medians around $12$-$14$\,dB, but Fourier and Haar exhibit a much broader upper tail, corresponding to the atlas-like images that are almost constant after percentile normalization. The grid is systematically shifted downwards, and SIREN has the tightest spread, consistent with its smooth, low-frequency bias.

\begin{figure}
    \centering
    \includegraphics[width=1\linewidth]{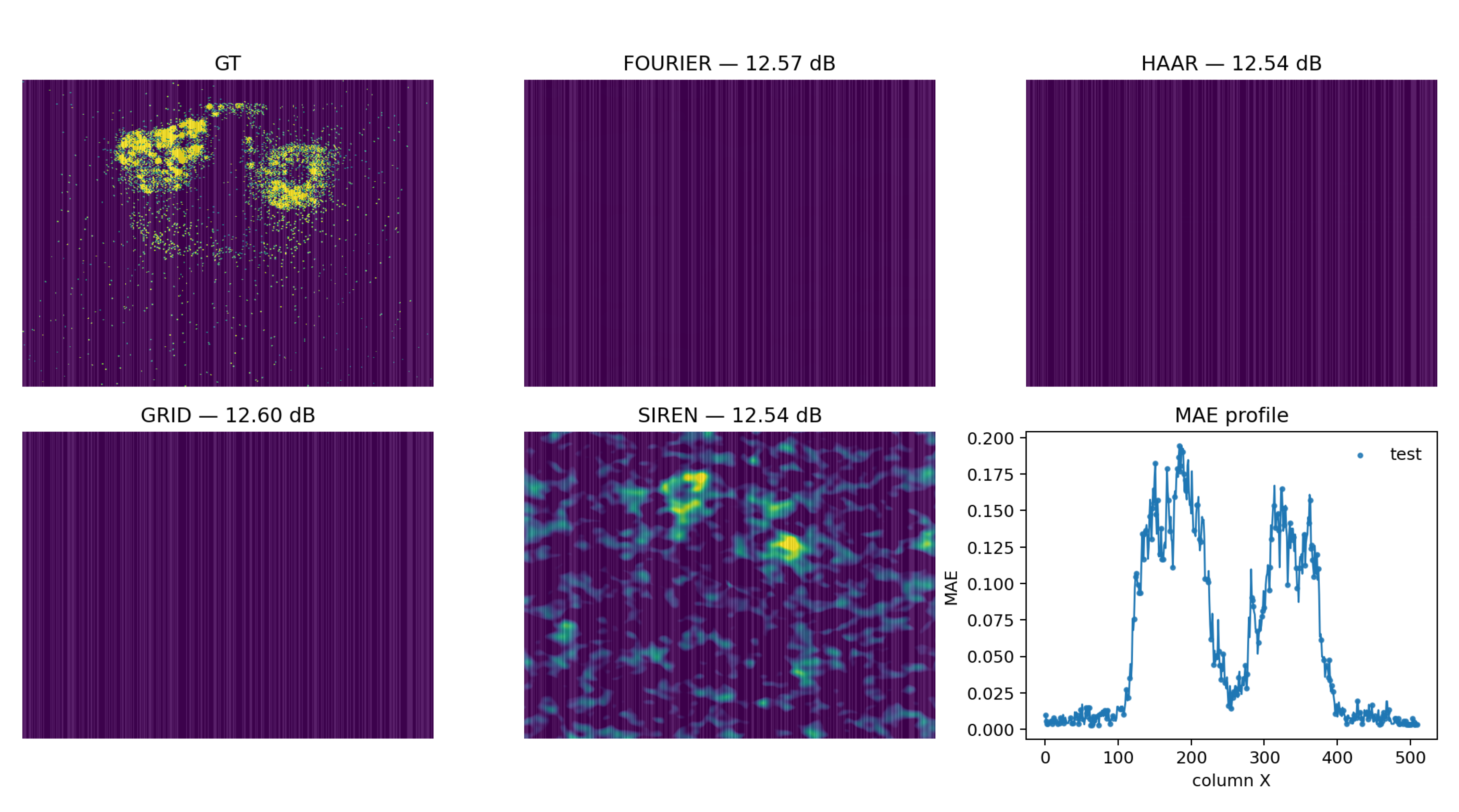}
    \caption{Representative reconstructions for selected MapZebrain regions using different positional encodings. Fourier and Haar preserve morphological detail and boundaries, SIREN smooths fine structures, and Grid exhibits strong blurring and vertical artifacts. Rightmost panels show column-wise MAE profiles.}
    \label{fig:figure_main_projection}
\end{figure}

Representative reconstructions for selected MapZebrain atlas regions illustrate clear qualitative differences between the INR parameterizations, see Figure \ref{fig:figure_main_projection}. Across both projection and column-wise holdout regimes, Fourier and Haar encodings preserve fine-grained textures, neuron-like puncta, and sharp regional borders, whereas SIREN produces globally smooth reconstructions that suppress small-scale structure. Grid exhibits strong vertical banding and loss of detail, indicating over-smoothing and poor generalization along the held-out axis. Haar’s multiscale basis successfully retains thin processes but occasionally introduces blocking artifacts at region boundaries-consistent with the discontinuities inherent in piecewise wavelet support. Fourier maintains overall coherence and edge continuity, yielding the most visually faithful reconstructions.

SIREN, relying purely on learned periodicity, captures low-frequency contrast but fails to recover localized intensity variations, while Grid’s fixed positional bins lead to severe blurring. The corresponding MAE profiles (right column) confirm these observations: Fourier and Haar maintain low, structured error across the spatial domain, while SIREN and Grid exhibit higher and more erratic MAE, especially in dense or highly textured regions.

Haar and Fourier consistently reach the highest $\mathrm{PSNR}_{\text{test}}$ and lowest MAE, while SIREN performs moderately and Grid remains weakest. Paired Wilcoxon signed-rank tests run separately for the two regimes confirmed that, in the all-in-one set, both Haar and Fourier significantly outperform the grid encoder ($p \ll 10^{-6}$, paired Cliff’s $\delta \approx 0.7$-$0.95$), and Haar also outperforms SIREN ($p < 10^{-5}$). In the regions set, differences between Haar, Fourier, and the grid are no longer significant ($p>0.4$), which is consistent with their almost identical distributions dominated by $\geq 100$\,dB outliers. Across both regimes, SIREN is the only method that never produces such extreme values.

Overall, Haar and Fourier emerge as the most robust INR encodings for MapZebrain data, providing a good balance between fidelity and computational cost, while Grid remains slightly behind in this low-epoch setting.

\begin{table}[h]
\centering
\caption{Per-regime results: macro-averaged $\mathrm{PSNR}_{\text{test\_mean}}$ (dB) and wins per method.}
\label{tab:macro_micro_regime}
\begin{tabular}{llcccc}
\toprule
Regime & Model & Macro-Mean & Macro-Std & Wins \\
\midrule
all-in-one & Haar    & 12.41 & 2.53 & 291 \\
all-in-one & Fourier & 12.93 & 1.79 & 198 \\
all-in-one & SIREN   & 12.70 & 0.82 & 228 \\
all-in-one & Grid    & 10.25 & 2.15 &   6 \\
\midrule
regions    & Haar    & 69.68 & 26.36 &  63 \\
regions    & Fourier & 67.43 & 26.22 &  90 \\
regions    & Grid    & 68.29 & 27.84 &  69 \\
regions    & SIREN   & 17.72 &  0.64 &   5 \\
\bottomrule
\end{tabular}

\end{table}

Table~\ref{tab:macro_micro_regime} reports macro- and micro-averaged $\mathrm{PSNR}_{\text{test\_mean}}$ across both regimes. In all-in-one (723 images), macro PSNR is $\approx$10-13\,dB (Fourier $12.93\pm1.79$, SIREN $12.70\pm0.82$, Haar $12.41\pm2.53$, Grid $10.25\pm2.15$), with wins: Haar $291$, SIREN $228$, Fourier $198$, Grid $6$. In regions (227 images), explicit encodings reach much higher macro PSNR (Haar $69.68\pm26.36$, Grid $68.29\pm27.84$, Fourier $67.43\pm26.22$, SIREN $17.72\pm0.64$), with wins: Fourier $90$, Grid $69$, Haar $63$, SIREN $5$. This split suggests that curated atlas slices favor explicit, multiscale/spectral encodings, while the heterogeneous all-in-one set narrows gaps and increases the relative competitiveness of SIREN on pixel-weighted micro averages.

To complement PSNR averages, we also counted wins per image, i.e. how many test images were best reconstructed by a given INR (highest $\mathrm{PSNR}_{\text{test}}$). In the heterogeneous {all-in-one set (723 images), Haar won on 291 images (40.20\%), SIREN on 228 (31.50\%), Fourier on 198 (27.40\%), and the grid only on 6 images (0.80\%). In contrast, in the curated regions set (227 images), explicit encodings dominated: Fourier won on 90 images (39.60\%), Haar on 63 (35.70\%), and the grid on 69 (22.50\%), while SIREN was best only on 5 images (2.20\%). These counts indicate that on clean, atlas-like slices Fourier/Haar are clearly preferable, on mixed-resolution, neuron-projection data SIREN is substantially more competitive than what its macro PSNR alone would suggest.
	
We note that very high PSNR values ($\geq 100$\,dB) occur only in the regions regime and correspond to nearly constant fields after per-image $(1,99)$ percentile normalization. Because such cases inflate macro means, we additionally report medians, IQRs, and 10\% trimmed means (Appendix, Table~\ref{tab:robu}). These robust statistics confirm the ranking Haar $\approx$ Fourier $>$ Grid $>$ SIREN.

Table~\ref{tab:ssim_edge} reports complementary test-only metrics (SSIM and edge MAE). Fourier attains the highest SSIM alongside the lowest edge error, indicating a favorable trade-off between global structure and boundary sharpness.
Grid shows SSIM close to Fourier yet exhibits markedly worse edge fidelity, consistent with smooth reconstructions that raise perceptual similarity but blur interfaces. SIREN reaches mid-rank SSIM and competitive PSNR\(_{\text{micro}}\), but degrades on edges relative to Fourier. Haar lies between Fourier and SIREN.

\begin{table}[ht]
	\centering
	\caption{Complementary test-only metrics on held-out columns (all images). Higher is better for SSIM; lower is better for Edge~MAE. PSNR$_\text{micro}$ reported for reference. Macro averages across images.}
	\label{tab:ssim_edge}
	\begin{tabular}{lccc}
		\toprule
		Model & SSIM$_\text{test}$ (↑) & Edge~MAE$_\text{test}$ (↓) & PSNR$_\text{micro}$ [dB] \\
		\midrule
		Fourier & \textbf{0.689} & \textbf{0.183} & \textbf{13.80} \\
		Grid    & 0.687          & 0.246          & 10.93 \\
		Haar    & 0.634          & 0.213          & 12.83 \\
		SIREN   & 0.443          & 0.208          & 13.43 \\
		\bottomrule
	\end{tabular}
\begin{minipage}{0.9\linewidth}
	\small
	\textit{Note.} All metrics are computed only on held-out columns; Edge~MAE is restricted to ground-truth edge pixels (top $10.00\%$ Sobel magnitudes per image).
\end{minipage}

\end{table}

We report test-only SSIM on the held-out columns using the standard $11\times 11$ Gaussian window after per-image $(1,99)$ normalization. Edge~MAE is the mean absolute error computed only over edge pixels detected in the ground truth by a Sobel operator, selecting the top $10.00\%$ of gradient magnitudes within the held-out columns. Fourier attains the best combination of global structure and boundary fidelity (highest SSIM and lowest Edge~MAE), which is most relevant for atlas registration. Grid's relatively high SSIM but poor Edge~MAE indicates over-smoothing of interfaces, while SIREN remains competitive in area-weighted PSNR but attenuates fine detail.

\section{Interpretation of results based on biological experiment}
On atlas slices, Fourier and Haar best preserve neuropil layering (e.g., stratified optic tectum) and sharp inter-regional borders (e.g., pallium-diencephalon interfaces, rhombomeric boundaries), which is consistent with their higher medians and larger share of per-image wins in the regions regime (Figure \ref{fig:box_overall_psnr}). In single-neuron projections within the all-in-one set, both encodings maintain slender process continuity across the column-wise held-out gaps and preserve commissural crossing at the midline (Figure \ref{fig:figure_main_projection}). By contrast, SIREN’s smooth bias attenuates high-spatial-frequency detail, leading to loss of the thinnest axonal and dendritic branches despite competitive area-weighted (micro) PSNR. The grid encoder occasionally introduces vertical banding aligned with the held-out axis, degrading boundary fidelity and commissural continuity. Taken together, these biology-aware observations align with known larval zebrafish morphology, layered neuropil, crisp region boundaries, and long, thin tracts, and explain why explicit spectral/multiscale encodings (Fourier/Haar) are preferable for atlas registration and boundary-sensitive tasks, while SIREN remains a lightweight baseline for background modeling or denoising where fine morphology is less critical.

\section{Discussion}
Tasks that hinge on boundary fidelity (e.g.\ region delineation, atlas registration) benefit most from explicit positional bases (Haar/Fourier), which better preserve inter‐regional edges. Tracing of sparse single, neuron projections favours encodings that maintain process continuity across missing spans; here Haar/Fourier were superior under the column‐wise split. SIREN’s smooth bias can still be attractive for background modelling or denoising in volumetric acquisitions where high‐frequency detail is less critical.

Our benchmark shown consistent differences between INR parameterizations. Explicit encodings (Fourier, Haar, and the multi-resolution  Grid) reach the best reconstruction quality, with Fourier/Haar leading on macro PSNR and Grid slightly behind under the short training budget. SIREN trails on macro but remains competitive on pixel-weighted micro averages in the heterogeneous all-in-one set. These trends align with prior work on frequency bias \cite{tancik2020,sitzmann2020}, where explicit positional bases mitigate the low-frequency preference of standard MLPs.

\begin{figure}
    \centering
    \includegraphics[width=1\linewidth]{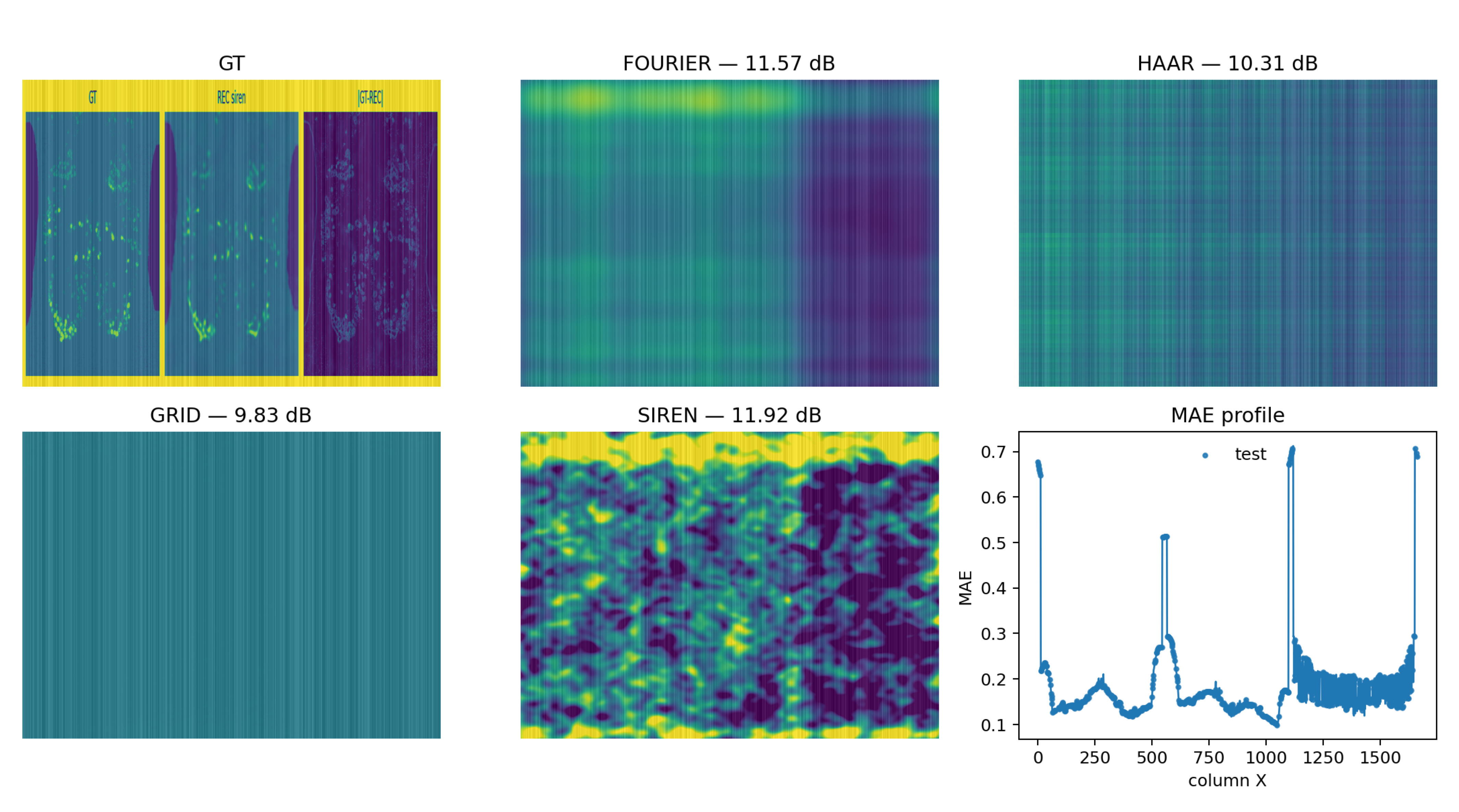}
    \caption{Smooth bias characteristic of sinusoidal INRs (SIREN).}
    \label{fig:siren_bias}
\end{figure}

Across both regimes, the relative ordering between parameterizations is stable and mirrors classic frequency-bias results, including explicit encodings that inject either dense spectral coverage (Fourier, Grid) or multiscale structure (Haar) consistently outperform purely learned periodicity (SIREN) on fine morphology. Smooth bias characteristic of sinusoidal INRs was illustraited in Figure \ref{fig:siren_bias}. This supports the view that, for neuroanatomical INR, the positional basis is a first-order design choice, while additional depth/capacity yields diminishing returns unless it aligns with the image statistics (thin processes, sharp boundaries) (\cite{sitzmann2020}).

The gap between macro- and micro-averaged scores indicates different failure modes: methods that excel in global contrast and edge sharpness (Fourier/Grid) may not maximize local smoothness, whereas SIREN, despite capturing low-frequency trends, tends to oversmooth small-scale textures. Example reconstruction under column-wise holdout was presented in Figure \ref{fig:columns}. For downstream biomedical tasks, this implies reporting both macro and micro metrics and selecting the parameterization by task: segmentation and region delineation benefit from edge-fidelity, while visualization or background modeling may tolerate smooth bias (\cite{rahaman2018}).

\begin{figure}
    \centering
    \includegraphics[width=1\linewidth]{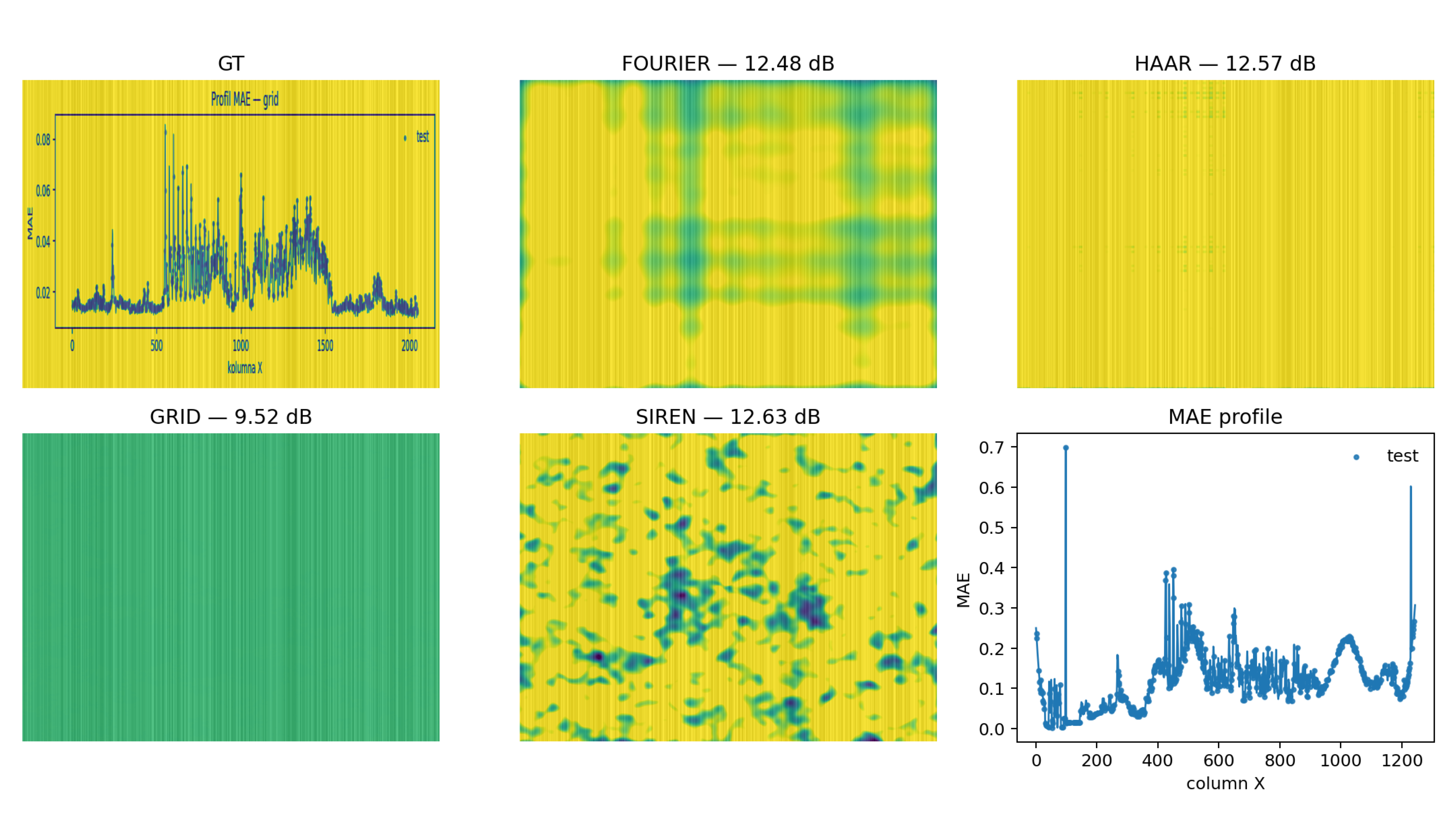}
    \caption{Example reconstruction under column-wise holdout regime illustrating directional overfitting.}
    \label{fig:columns}
\end{figure}

The column-wise hold-out emphasizes interpolation across anisotropic missing spans rather than memorization. Methods with frequency-structured priors (Fourier/Grid) maintain fidelity under this split, suggesting improved out-of-grid generalization, which is relevant for mosaics, partial acquisitions, and sparsely sampled volumes commonly encountered in microscopy pipelines (\cite{tancik2020}).

\begin{figure}
    \centering
    \includegraphics[width=1\linewidth]{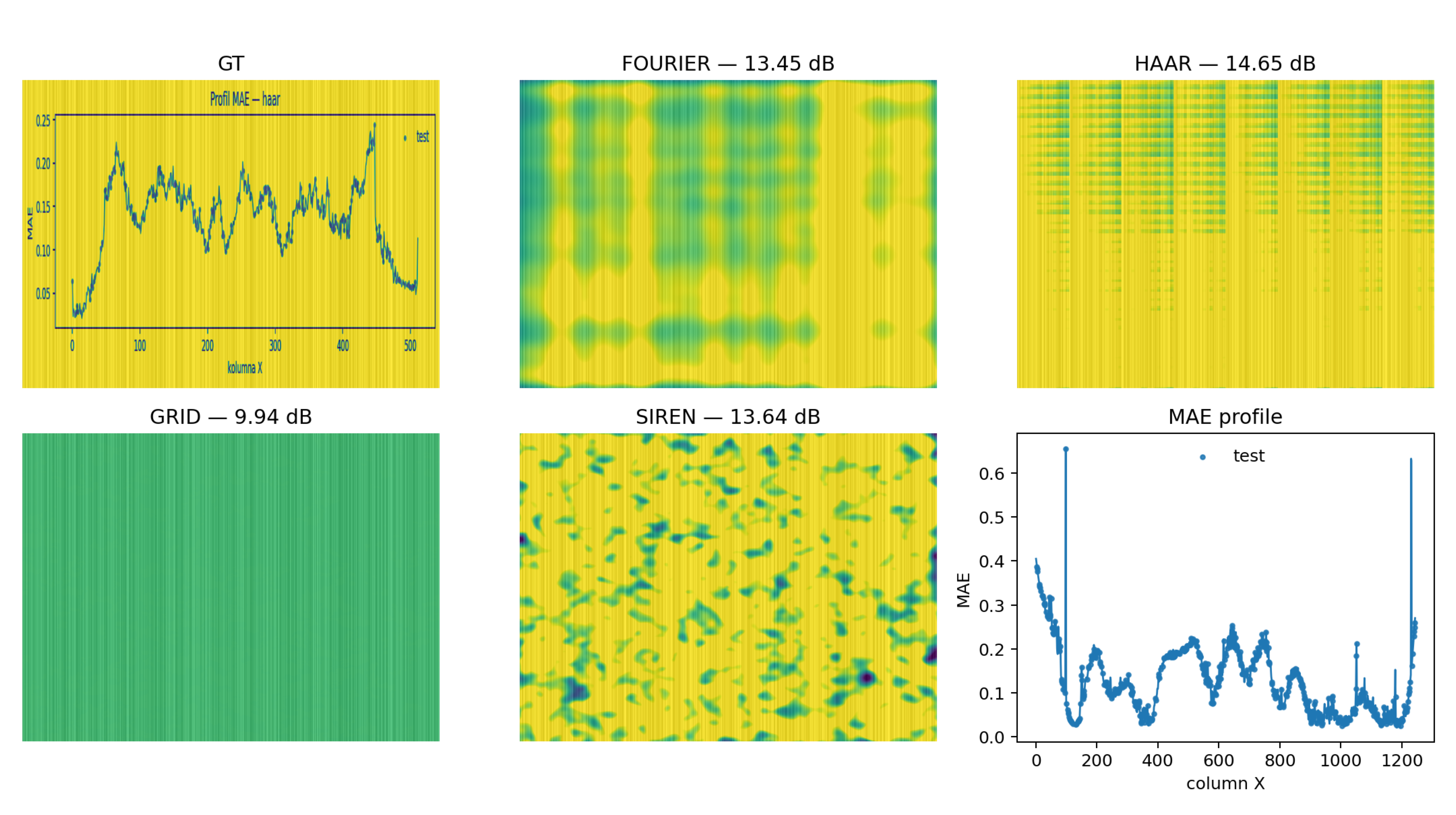}
    \caption{Haar positional encoding preserves sharp edges but introduces block boundary artifacts.}
    \label{fig:haar_artifacts}
\end{figure}

Haar’s blocking artifacts at region boundaries point to a basis-data mismatch at block transitions (\cite{tancik2020}). Haar positional encoding preserves sharp edges but introduces block boundary artifacts (Figure~\ref{fig:haar_artifacts}). Haar achieves high fidelity ($\approx$ 14.6 dB PSNR) and retains fine processes, introduces soft boundary seams between neighboring blocks, consistent with the piecewise formulation of Haar encodings. Fourier yields smoother transitions but blurs thin structures, while SIREN and Grid again show excessive smoothing. The MAE profile reveals localized oscillations aligned with block boundaries. These artifacts can be mitigated by overlapping windows or soft wavelet variants, which preserve edge faithfulness while improving continuity. Soft multiscale variants, like overcomplete or learned wavelet dictionaries, could mitigate this while retaining edge faithfulness (\cite{mueller2022}). 

For boundary-sensitive downstream tasks (atlas registration, label transfer), prefer Fourier or Haar. For background modeling or denoising under tight compute, SIREN is a viable baseline. Grid encoders may require longer training and tuning (levels, features per level) to reach their potential on microscopy textures.

\section{Practical implications for atlas workflows}
For readers using larval zebrafish atlases in practice, our results suggest simple guidelines. First, for boundary-sensitive tasks such as registration, label transfer, and region-of-interest analysis on MapZebrain-like data, Fourier or Haar positional encodings should be preferred, as they consistently preserve inter-regional borders and layered neuropil while maintaining high edge fidelity. Second, for sharing single-neuron projections in a storage-efficient way, both Fourier and Haar achieve a favourable trade-off between reconstruction quality and model size, retaining thin commissural processes and local punctate texture. Third, SIREN provides a useful
baseline when computational resources are limited or when the primary goal is
background modelling or smoothing, but it should be used with caution when fine
morphology is critical. Finally, our column-wise hold-out protocol can be adopted as a routine sanity check for assessing whether new INR variants genuinely generalize across contiguous gaps typical of mosaics and sparse-view acquisitions.

\section{Limitations}
Our study applies a short training budget for both regimes, namely 10.00\% pixel sampling, 10 epochs. Longer optimization or hyperparameter tuning could change the relative ranking, particularly for Grid encoders. We train models independently for each image, without using cross-image priors or transfer. Consequently, cross-specimen generalization is not evaluated. Reported PSNR distributions are broad, reflecting heterogeneous native resolutions and sparsity. Very high outliers, equal or higher than 100\,dB, typically occur for nearly constant or small fields of view. 

\section{Conclusions}
We introduced a two-regime, reproducible INR benchmark on MapZebrain spanning region-level atlas slices and whole-projection images under a unified protocol. Haar and Fourier positional encodings consistently achieved the best reconstruction fidelity (macro-averaged $\mathrm{PSNR}_{\text{test\_mean}}\!\approx\!26\,\mathrm{dB}$), with Grid performing moderately and SIREN remaining competitive mainly on micro-averaged metrics. Across both regimes, multiscale (Haar) and dense-spectral (Fourier) priors preserved thin processes and sharp boundaries substantially better than architectures relying on learned smoothness (SIREN, Grid), indicating that the inductive bias of the positional basis outweighs marginal depth/capacity gains. The systematic gap between macro- and micro-averaged scores shows that methods trade off global boundary integrity versus local smoothness, underscoring the value of reporting both.
Our column-wise holdout emphasizes interpolation rather than memorization, suggesting that frequency-structured priors generalize more reliably to out-of-grid coordinates typical in partially sampled biomedical imagery.

\begin{table}[h]
	\centering
	\caption{Key hyperparameters}
	\label{tab:impl_hparams}
	\begin{tabular}{l l}
		\toprule
		Setting & Value \\
		\midrule
		Intensity normalization & per-image percentiles $(1,99)$ \\
		Hold-out split & \texttt{blocked\_cols\_X}, $\alpha=0.40$ \\
		Train subsampling & $10\%$ pixels from training columns \\
		Epochs / Optimizer & $10$ / AdamW ($\mathrm{lr}=10^{-3}$, $\mathrm{wd}=10^{-6}$) \\
		Loss & Smooth-$\ell_1$ ($\beta=0.01$) \\
		Batch size & $\min(131{,}072,\; \max(65{,}536,\; \lfloor HW\cdot0.10/2 \rfloor))$ \\
		Dataloading & \texttt{num\_workers}=0, \texttt{pin\_memory}=True \\
		Device & CPU (CUDA only if available; not used here) \\
		SIREN & width=256, depth=6, $w_0^{\text{first}}=36$, $w_0^{\text{hidden}}=1$ \\
		Fourier & bands=48, max\_freq=24, learnable, + MLP(192, 4) \\
		Haar & L=8, include\_input, + MLP(192, 4) \\
		Grid & L=8, feats=2, + MLP(192, 4) \\
		\bottomrule
	\end{tabular}
\end{table}

\section{Reproducibility}
All experiments ran deterministically on an Intel(R) Core(TM) i7-14700F (2.10\,GHz), Windows 10 (x86\_64), Python 3.10.13; PyTorch 2.1.2+cpu, torchvision 0.16.2+cpu, torchaudio 2.1.2+cpu; NumPy 1.26.4 (OpenBLAS), SciPy 1.11.4, pandas 2.1.4, Matplotlib 3.8.2, scikit-image 0.22.0, PyYAML 6.0.1, tqdm 4.66.2. Global seeding uses \texttt{numpy.random.default\_rng(SEED\_GLOBAL)}; per-file test masks are derived deterministically from the file path (stable hash) combined with \texttt{SEED\_GLOBAL}. Train/test splits and pixel subsampling are fixed across methods.

All models receive $(x,y)\!\in\![0,1]^2$ coordinates and predict a single grayscale intensity. Each image is converted to grayscale and independently normalized by robust percentiles $(1,99)$:
\[
I_{\mathrm{norm}} \;=\; \mathrm{clip}\!\left(\frac{I - P_1}{P_{99} - P_1 + 10^{-6}},\,0,\,1\right).
\]
We train on a column-wise hold-out along the $X$-axis with a fixed test fraction $\alpha=0.40$ (\texttt{HOLDOUT\_FRAC\_X=0.40}), and randomly subsample $10.00\%$ of pixels \emph{only} from training columns (\texttt{TRAIN\_SAMPLE\_PERC=0.10}). No resizing is applied in the \emph{regions} regime (\texttt{RESIZE\_TO=None}). Optimization uses AdamW with a fixed learning rate $10^{-3}$ and weight decay $10^{-6}$ for 10 epochs (\texttt{EPOCHS=10}). The loss is Smooth-$\ell_1$ with $\beta=0.01$. The batch size is computed adaptively as
\[
\texttt{bs} \;=\; \min(\texttt{BATCH\_CAP}=131{,}072,\; \max(65{,}536,\; \lfloor (H\cdot W \cdot 0.10)/2 \rfloor)).
\]
Data loaders use \texttt{num\_workers=0} and \texttt{pin\_memory=True}. Training runs on CPU (CUDA is used only if available; in our experiments CUDA was not used). Key parameters are shown in the Table \ref{tab:impl_hparams}.
Train/test splits and random subsampling are fully deterministic (\texttt{numpy.default\_rng(seed)}) and identical across INR variants.

\section*{Appendix}
\subsection*{Detailed summary statistics}
Table~\ref{tab:overall_stats} reports descriptive statistics of per-image test PSNR (macro view) across all 950 images. Table~\ref{tab:regime_stats} breaks these down by regime.

\begin{table}[ht]
	\centering
	\caption{Per-method descriptive statistics of $\mathrm{PSNR}_{\text{test\_mean}}$ (dB) across all images (N=950): mean$\pm$SD, median, and IQR.}
	\label{tab:overall_stats}
	\begin{tabular}{lcccc}
		\toprule
		Model & Mean $\pm$ SD & Median & IQR & N \\
		\midrule
		Fourier & 25.95 ± 26.59 & 13.16 & [11.87, 18.05] & 950 \\
		Grid & 24.12 ± 28.31 & 9.99 & [9.36, 16.43] & 950 \\
		Haar & 26.09 ± 27.70 & 14.10 & [10.31, 16.92] & 950 \\
		SIREN & 13.90 ± 2.28 & 13.00 & [12.21, 15.18] & 950 \\
		\bottomrule
	\end{tabular}
\end{table}

\begin{table}[ht]
	\centering
	\caption{Per-method descriptive statistics by regime (macro view): mean$\pm$SD, median, and IQR.}
	\label{tab:regime_stats}
	\begin{tabular}{l l c c c c}
		\toprule
		Regime & Model & Mean $\pm$ SD & Median & IQR & N \\
		\midrule
		all-in-one & Fourier & 12.93 ± 1.79 & 12.47 & [11.73, 13.69] & 723 \\
		all-in-one & Grid & 10.25 ± 2.15 & 9.80 & [9.06, 10.32] & 723 \\
		all-in-one & Haar & 12.41 ± 2.53 & 12.96 & [10.18, 14.52] & 723 \\
		all-in-one & SIREN & 12.70 ± 0.82 & 12.54 & [12.08, 13.23] & 723 \\
		regions & Fourier & 67.43 ± 26.22 & 55.81 & [45.77, 100.00] & 227 \\
		regions & Grid & 68.29 ± 27.84 & 54.89 & [44.36, 100.00] & 227 \\
		regions & Haar & 69.68 ± 26.36 & 64.10 & [46.50, 100.00] & 227 \\
		regions & SIREN & 17.72 ± 0.64 & 17.74 & [17.38, 18.17] & 227 \\
		\bottomrule
	\end{tabular}
\end{table}

\subsection*{Pairwise significance (macro PSNR)}
We report paired Wilcoxon signed-rank tests (two-sided) on per-image PSNR with Holm-Bonferroni correction in the main paper, see Table \ref{tab:pairwise}. Here we additionally provide Cliff's $\delta$ effect sizes with bootstrap 95\% CIs (400 resamples) and the number of paired images used.

\begin{table}[ht]
	\centering
	\caption{Pairwise comparisons: macro PSNR.}
	\label{tab:pairwise}
	\begin{tabular}{lcccc}
		\toprule
		Comparison & $p$-value (Wilcoxon) & Cliff's $\delta$ & 95\% CI & Paired N \\
		\midrule
		Haar vs. SIREN & 2.617e-14 & 0.051 & [-0.008, 0.103] & 950 \\
		Fourier vs. SIREN & 2.940e-34 & 0.029 & [-0.020, 0.079] & 950 \\
		Haar vs. Grid & 1.692e-49 & 0.308 & [0.254, 0.359] & 950 \\
		Fourier vs. Grid & 5.293e-67 & 0.435 & [0.387, 0.487] & 950 \\
		Haar vs. Fourier & 1.432e-08 & -0.053 & [-0.107, 0.001] & 950 \\
		\bottomrule
	\end{tabular}
\end{table}

\subsection*{Outlier analysis (regions, $\ge 100$ dB)}
High PSNR outliers occur only in the regions set and correspond to near-constant fields after per-image percentile normalization, see Table \ref{tab:outliers}.

\begin{table}[ht]
	\centering
	\caption{Outliers (PSNR $\ge 100$ dB) in \emph{regions}.}
	\label{tab:outliers}
	\begin{tabular}{lccc}
		\toprule
		Model & Outliers & Total (regions) & Share (\%) \\
		\midrule
		Fourier & 73 & 227 & 32.2 \\
		Haar    & 67 & 227 & 29.5 \\
		Grid    & 94 & 227 & 41.4 \\
		SIREN   & 0  & 227 & 0.0 \\
		\bottomrule
	\end{tabular}
\end{table}

\subsection*{Wins per regime}
Counts and shares (percentage within regime) of images won by each method, see Tables \ref{tab:wins_regime}, \ref{fig:wins_all_in_one}, and \ref{fig:wins_regions}.

\begin{table}[ht]
	\centering
	\caption{Wins per regime with shares.}
	\label{tab:wins_regime}
	\begin{tabular}{l l c c c}
		\toprule
		Regime & Best method & Wins & N & Share (\%) \\
		\midrule
		all-in-one & Fourier & 198 & 723 & 27.4 \\
		all-in-one & Grid & 6 & 723 & 0.8 \\
		all-in-one & Haar & 291 & 723 & 40.2 \\
		all-in-one & SIREN & 228 & 723 & 31.5 \\
		regions & Fourier & 90 & 227 & 39.6 \\
		regions & Grid & 69 & 227 & 22.5 \\
		regions & Haar & 63 & 227 & 35.7 \\
		regions & SIREN & 5 & 227 & 2.2 \\
		\bottomrule
	\end{tabular}
\end{table}

\begin{figure}[h]
	\centering
	\includegraphics[width=0.75\linewidth]{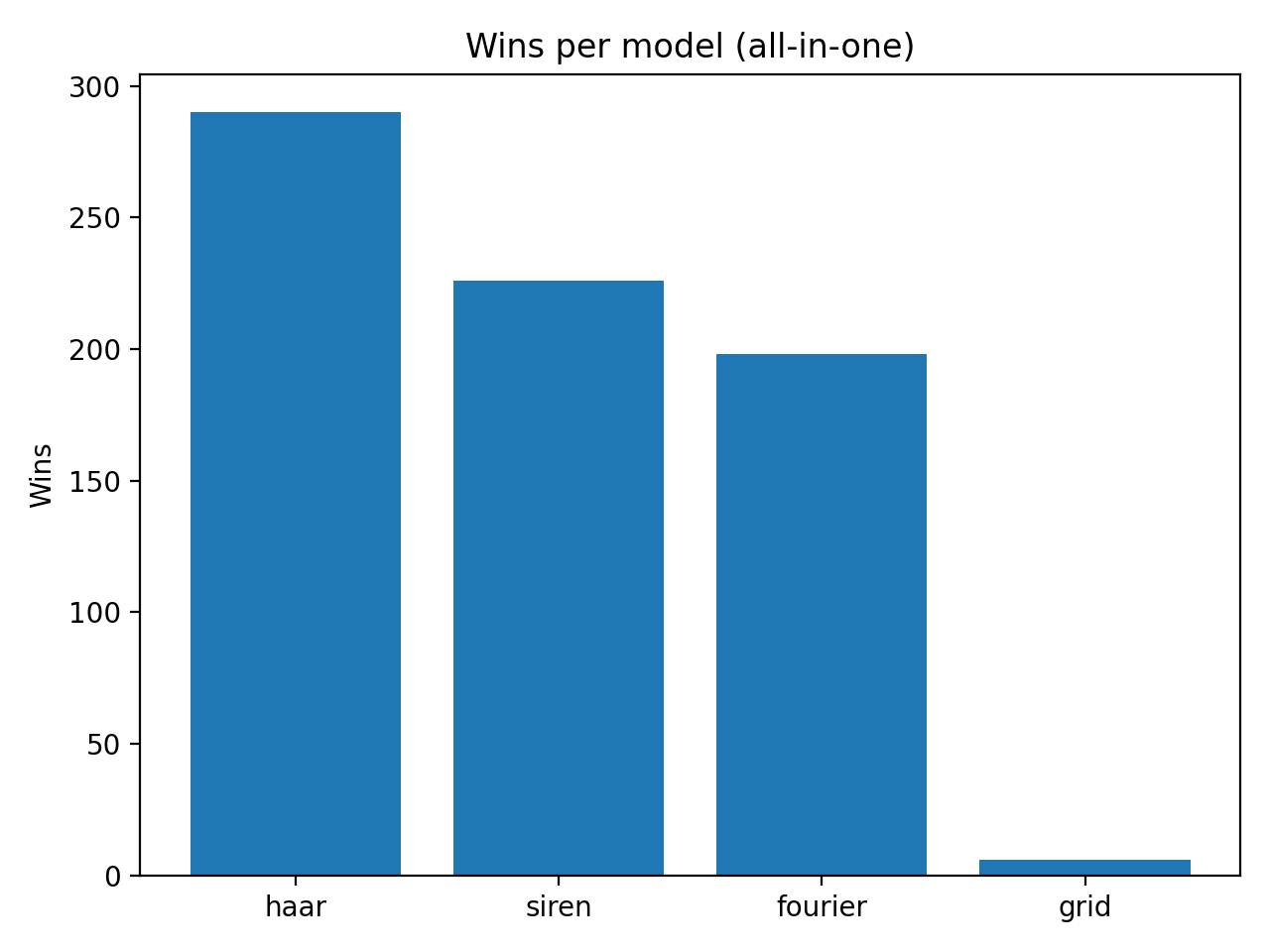}
	\caption{Number of per-image wins in the \textit{all-in-one} regime. Haar dominates, SIREN remains competitive, the grid almost never wins.}
	\label{fig:wins_all_in_one}
\end{figure}

\begin{figure}[h]
	\centering
	\includegraphics[width=0.75\linewidth]{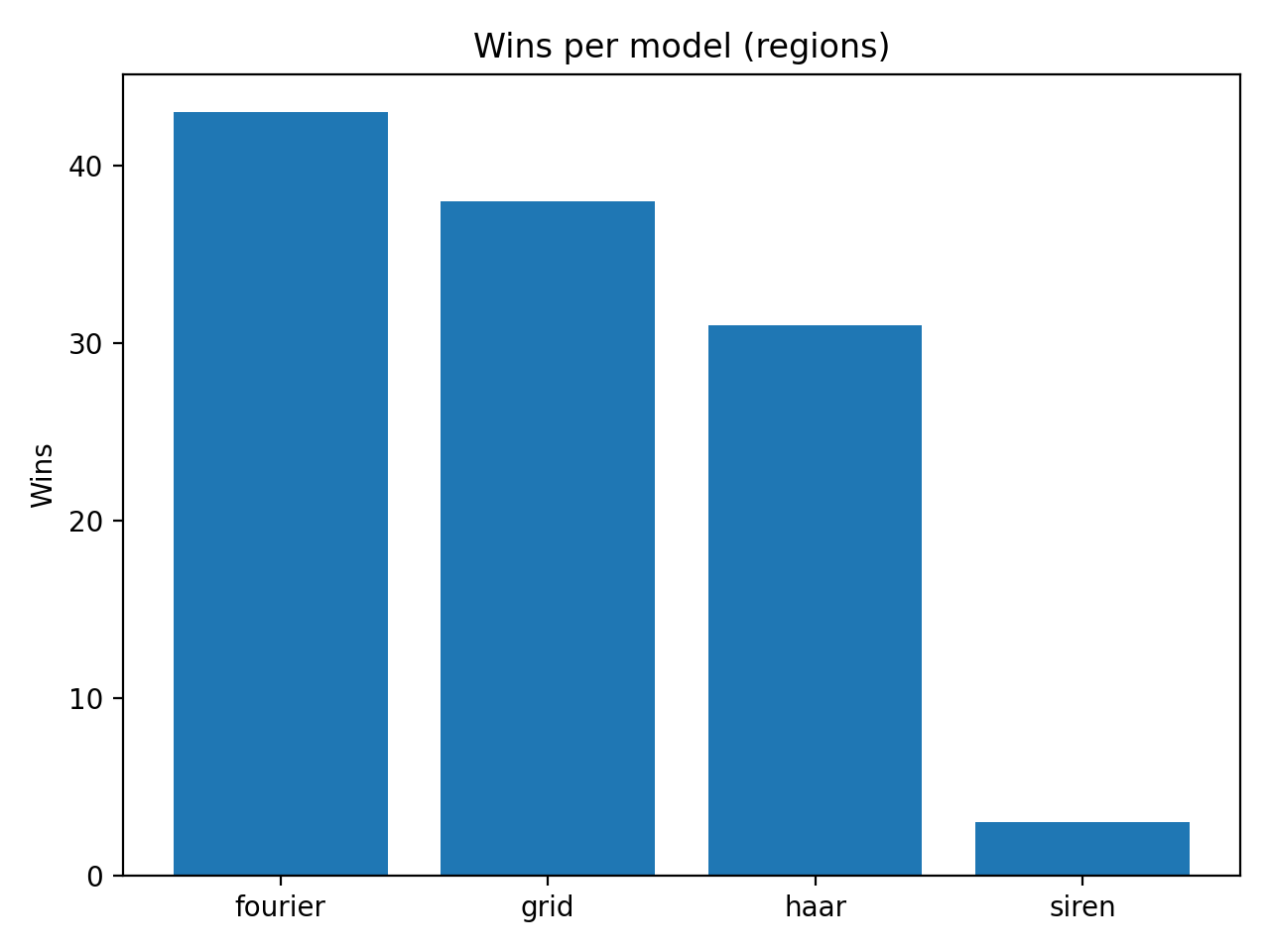}
	\caption{Number of per-image wins in the \textit{regions} regime. Explicit encodings (Fourier, Haar, Grid) dominate over SIREN.}
	\label{fig:wins_regions}
\end{figure}

\subsection*{Robust reporting}

In addition to means $\pm$ SD, we report median, IQR, and 10.00\% trimmed means to reduce the influence of near-constant outliers \ref{tab:outliers}, in Table \ref{tab:robu}.

\begin{table}[ht]
	\centering
	\caption{Robust PSNR statistics across all images.}
	\begin{tabular}{lccc}
		\toprule
		Model & Median & IQR & Trimmed Mean (10\%) \\
		\midrule
		Haar    & 14.10 & [10.31, 16.92] & 16.84 \\
		Fourier & 13.16 & [11.87, 18.05] & 16.27 \\
		Grid    &  9.99 & [ 9.36, 16.43] & 14.22 \\
		SIREN   & 13.00 & [12.21, 15.18] & 13.62 \\
		\bottomrule
	\end{tabular}
	\label{tab:robu}
\end{table}

\section*{Ethics Statement}
This study performs secondary analysis of publicly available microscopy data (MapZebrain) and involves no new experiments on animals or humans. No identifiable human data are present; therefore institutional ethics approval and consent were not required.

\section*{Conflict of interest}
The author declares no competing interests.

\section*{Data availability}
All raw data are available at \cite{mapzebrain_site}. Preprocessed splits, masks, and scripts to reproduce the experiments will be made available on request.

\section*{Funding}
This research received no external funding.

\section*{Author contribution}
AP: Conceptualization, Methodology, Software, Validation, Formal analysis, Investigation, Data curation, Visualization, Writing-original draft, Writing-review \& editing.

\FloatBarrier    

\end{document}